%% file: emnlp2023.tex
\pdfoutput=1
\documentclass[11pt]{article}

\usepackage[]{EMNLP2023}

\usepackage{times}
\usepackage{latexsym}
\usepackage{graphicx}
\usepackage[T1]{fontenc}
\usepackage[utf8]{inputenc}
\usepackage{microtype}
\usepackage{algorithm}
\usepackage{algpseudocode}
\usepackage{inconsolata}

\usepackage{microtype}
\usepackage{graphicx}
\usepackage{multirow}
\usepackage{multicol}
\usepackage{microtype}
\usepackage{subcaption}
\usepackage{enumitem,kantlipsum}
\usepackage[normalem]{ulem}
\usepackage{tabularx}
\usepackage{tcolorbox}
\usepackage{color, colortbl}
\usepackage{todonotes} 

\usepackage{pbox}
\usepackage{booktabs} 
\usepackage{amssymb}% http://ctan.org/pkg/amssymb
\usepackage{pifont}% http://ctan.org/pkg/pifont

\usepackage{tabu}

\usepackage{fontawesome}
\usepackage{algorithm} 
\usepackage{algpseudocode}
\usepackage{amsmath}

\title{Pseudo-Labeling for Domain-Agnostic \\Bangla Automatic Speech Recognition}
\author{Rabindra Nath Nandi$^1$, Mehadi Hasan Menon$^1$, Tareq Al Muntasir$^1$, \\ {\bf Sagor Sarker$^1$, Quazi Sarwar Muhtaseem$^1$, Md. Tariqul Islam$^1$,} \\
{\bf Shammur Absar Chowdhury$^2$, Firoj Alam$^2$}\\
  $^1$Hishab Singapore Pte. Ltd, Singapore \\
  $^2$Qatar Computing Research Institute, HBKU, Doha, Qatar \\
  \texttt{rabindra.nandi@hishab.co, shchowdhury@hbku.edu.qa} 
  \\}
\begin{document}
\maketitle

\begin{abstract}
One of the major challenges for developing automatic speech recognition (ASR) for low-resource languages is the limited access to labeled data with domain-specific variations. In this study, we propose a pseudo-labeling approach to develop a large-scale domain-agnostic ASR dataset. With the proposed methodology, we developed a 20k+ hours labeled Bangla speech dataset covering diverse topics, speaking styles, dialects, noisy environments, and conversational scenarios. We then exploited the developed corpus to design a conformer-based ASR system. We benchmarked the trained ASR with publicly available datasets and compared it with other available models. To investigate the efficacy, we designed and developed a human-annotated domain-agnostic test set composed of news, telephony, and conversational data among others. Our results demonstrate the efficacy of the model trained on psuedo-label data for the designed test-set along with publicly-available Bangla datasets. The experimental resources will be publicly available.\footnote{\url{https://github.com/hishab-nlp/Pseudo-Labeling-for-Domain-Agnostic-Bangla-ASR}}
\end{abstract}

\input{sections/introduction}
\input{sections/related_work}

\input{sections/dataset}

\input{sections/experiments}

\input{sections/results_and_discussion}

\section{Conclusion and Future Work}
\label{sec:conclusion}
This study offers a significant contribution in Bangla speech processing, in addition to the field of ASR particularly for low-resource language. The primary contribution of this paper lies in demonstrating that the model trained with pseudo-labeling only, offers comparable performance with supervised ASR systems. Specifically, the MegaBNSpeech model excels in their ability to generalize across multiple domains and channels as shown in the results.   

Additionally, the developed train, development, and two test sets of MegaBNSpeech corpus of $\approx 20,000$ hours of data will serve as a valuable resource for the research community. The MegaBNSpeech corpus, especially the manually annotated YT and telephony test sets, can be used as a benchmark for future studies, enabling other researchers to build upon our work and potentially discover even more effective methods for designing low-resource ASR.

% offers a significant contribution to the field of Bangla ASR research.  Its large-scale nature, comprising 18,000 hours of training data, provides a substantial resource for training state-of-the-art ASR models. Additionally, the provided test sets can be used for evaluating and benchmarking new models. Our pseud-labeling annotation strategy provides an important direction to be dependent on machine-generated labeling in case of large unlabeled datasets. 

\section*{Acknowledgments}
We are grateful to HISHAB\footnote{\url{https://hishab.co/}} for providing us with all the necessary working facilities, computational resources, and an appropriate environment throughout our entire work.

\section{Limitations}
Our data collection originated from YouTube and in-house telephony conversations. Due to restrictions on sharing most of the YouTube content directly, we will instead release links to the YouTube videos along with their transcriptions.

% Furthermore, the inclusion of 1,800 hours of carefully curated test data enables comprehensive evaluation and benchmarking of ASR. 

% Later 

\bibliography{bib/asr}
\bibliographystyle{acl_natbib}

% \appendix
% \clearpage
% \input{sections/appendix}

\end{document}

%% file: sections/introduction.tex
\section{Introduction}
\label{sec:introduction}

Modern end-to-end automatic speech recognition (E2E-ASR) systems have made remarkable strides, performing well across various types of data \cite{li2020developing, gulati2020conformer,chowdhury2021onemodel,prabhavalkar2023end}. This success can be attributed to the advancement of deep learning techniques relying on different training strategies, highly dependent on large datasets.
However, acquiring and maintaining these high-quality human transcriptions is both expensive and time-consuming, and hence hinders further progress for ASR especially in low-resource languages like Bangla.

To overcome these challenges, two dominant methods, leveraging unlabeled audio, are gaining popularity. These methods include: \textit{(i)} 
% unsupervised pre-training via 
pre-training via Self-supervised learning (SSL) \cite{baevski2020wav2vec,baevski2022data2vec,chung2021w2v,hsu2021hubert}; \textit{(ii)} pseudo-labeling (PL) \cite{kahn2020self, xu2020slimipl, manohar2021kaizen,zhu2023alternative,xu2020iterative,higuchi2022momentum}. In the pre-training approach, the model is initially trained on raw unlabeled data and then fine-tuned using limited labeled data for some downstream ASR tasks. In pseudo-labeling, a pre-trained model generates labels for unlabeled data, which are then used alongside real labels for supervised ASR training. This paradigm is widely adopted due to its simplicity and effectiveness. Both SSL and PL have been shown to achieve competitive results with minimal labeled data, hence making these paradigms, especially PL, suitable for low-resource languages.
 
% Both these methods achieve competitive results with minimal labeled data. Nonetheless, each approach have its own drawbacks: pre-training and then fine-tuning paradigm can be computationally expensive \cite{}, and pseudo-labeling always introduce errors in the dataset, the quality highly depends on the performance of the expert/teacher system \cite{}, noise filtering techniques \cite{} and prediction ensemble \cite{} methods.

Despite being the 6$^{th}$ most widely spoken language globally, Bangla still falls under low resource language family mainly due to the lack of accessible open %speech 
datasets. To reduce this gap, we introduce a pseudo-labeling approach to develop an extensive, large-scale, and high-quality speech dataset of $\approx 20,000$ hours for developing domain-agnostic Bangla ASR. 
First, we curated and cleaned the largest collection of Bangla audio-video data from various Bangla TV channels on YouTube (YT) -- varying domains, speaking styles, dialects, and communication channels among others.
% developed a robust data collection pipeline that systematically extracted audio segments from these channels, ensuring a wide coverage of speech samples.
We then leverage the alignments from two ASR systems, to segment and automatically annotate the audio segments. We enrich the quality of pseudo-labels with our confidence and duration-based filtering method. 
% and enriched with multiple filtering criterion, to segment and automatically annotate the audio segments. 
We utilize the created dataset to design an end-to-end state-of-the-art Bangla ASR. Finally, we benchmark the ASR with widely used, domain-agnostic test sets and compare it with both publicly and commercially available Bangla ASR systems. To test domain-generalization capability, we also developed manually annotated test sets that include domain-diverse speech segments. 

\noindent Our contributions are as follows:
\begin{itemize} [leftmargin=*]%[noitemsep,topsep=0pt]
\item We develop and release \textbf{MegaBNSpeech} -- the largest Bangla speech ($ \approx 20,000$ hours) training corpus, alongside with its metadata;
\item We introduce a robust data collection pipeline that systematically extracted audio segments from listed channels, ensuring wide coverage of speech samples;
\item We developed and publicly released a domain-agnostic state-of-the-art Bangla ASR model;
\item We developed two test sets comprising (a) diversified domain data from YT; and (b) real-life telephony conversational data, to test model generalizability across domains;
\item We benchmark the proposed domain-agnostic Bangla ASR with publicly available test data and ASR models.
\end{itemize}

\begin{table*}
\centering
\begin{tabular}{llll}
\hline
\textbf{Datasets} &\textbf{Duration (Hours)} &\textbf{Source} &\textbf{Annotation} \\
\hline
Fleurs \cite{conneau2023fleurs} & 15.61 & Wikipedia & Human \\
Common Voice13 \cite{ardila2019common}  & 65.71 & Open domain & Human \\
OpenSLR \cite{kjartansson2018crowd} & 229 & Open domain & Human \\
Bengali Speech Corpus \cite{ahmed2020preparation} & 960 & Youtube & Pseudo \\

OOD-Speech \cite{rakib2023ood} &12K &Open domain & Human \\ \midrule
\textbf{MegaBNSpeech (Ours)} & \textbf{19.8K} & YouTube & Pseudo \\
\hline\end{tabular}
\caption{A comparison of commonly used Bangla ASR datasets}
\label{tab:datasets_comparison}
\end{table*}

The rest of the paper is organized as follows: Section~\ref{sec:related_work} presents previous work, Section~\ref{sec:dataset} describes the dataset, Section~\ref{sec:experiments} formulates our experiments, Section~\ref{sec:results} discusses the evaluation results. Finally, Section~\ref{sec:conclusion} concludes and points to possible directions for future work.

%% file: sections/related_work.tex
\section{Related Work}
\label{sec:related_work}

% \todo[inline]{TO DO}

\subsection{Speech Datasets Development}
In the realm of speech corpus development, a variety of methods and techniques have been employed across multiple languages. For example, \citet{wang2005matbn} focused on Mandarin Chinese, creating a speech corpus from broadcast news and aligning the transcriptions. Similarly, \citet{radeck2015open} curated data from diverse sources like audiobooks and web recordings to create a comprehensive speech corpus for German. In terms of automatic speech recognition datasets, \citet{chui2008nccu} employed a method that constructs a Mandarin Chinese speech corpus using online videos and automated transcription. In a similar vein, \citet{cho2021kosp2e} harnessed web data and automatic alignment techniques to develop a Korean speech corpus geared toward speech recognition research.

Furthermore, current literature has also focused on specialized domains or applications. For instance, in the medical field, \citet{cho2021kosp2e} crafted a targeted speech corpus designed for medical dictation tasks, featuring recordings from healthcare professionals. Similarly, in the context of voice assistants, \citet{gale2019improving} developed a corpus explicitly aimed at training and evaluating voice-controlled systems.

\subsection{Speech datasets for Bangla}
There have been several recent works for Bangla Speech Recognition. \citet{sumit2018noise} proposed a deep learning based on approach and evaluated model on clean \cite{alam2010development} and noisy speech datasets \cite{bills2016iarpa}.
% In the field of Bangla, Automatic speech recognition Sakhawat Hosain Sumit is one of them who as well as other four authors have published a research article based on Noise Robust End-to-End Bangla Speech Recognition \cite{sumit2018noise}. 
% Shafayat Ahmed is one of them who prepared a Bangla speech corpus 
\citet{ahmed2020preparation} developed a large annotated speech corpus comprising 960 hours, which are automatically curated from publicly accessible audio and text data. 
% In the paper \cite{ahmed2020preparation} 
The data annotation primarily relies on publicly available audiobooks and TV news recordings from YouTube. It applies automated techniques such as format conversion, noise reduction, speaker diarization, and automatic gender detection. Transcriptions are generated iteratively using two speech recognition systems, with consensus determining accurate transcriptions. The resulting corpus, referred to as the `Transcribed corpus', encompasses approximately 510 hours of data.

% Bangla Speech Corpus consisting of 960 hours,  is prepared by using Bangla audiobook recordings and TV news from YouTube, and the annotation is weakly labeled. The authors \cite{ahmed2020preparation} used a pseudo labeling annotation pipeline. However, the annotation algorithm is quite different from our annotation pipeline. 

Similarly, \citet{rakib2023ood} created another extensive dataset with a focus on out-of-domain distribution generalization. The dataset is collected via crowdsourcing campaigns on the duration between Feb 2022 and Nov 2022 on the Mozilla Common Voice (MCV) platform. They followed two collection strategies: (i) scripted and (ii)spontaneous. The dataset contains 11.8k hours of training data curated from 22, 645 native Bangla speakers from South Asia. So far, this is the largest dataset available online for Bangla ASR Recognition. 
\citet{kibria2022bangladeshi} also developed a speech corpus that includes 241 hours of both recorded and broadcast speech, featuring contributions from over 60 speakers.
% Fazle Rabbi Rakib is another researcher who as well as his team has worked with a Large Bangla Speech Recognition Dataset for OOD-Speech \cite{rakib2023ood}. 
% Mohammad Shahidur Rahman, another researcher, Has worked on the Bangla speech corpus as well. \cite{kibria2022bangladeshi}. 
%safi 

Fleur's datasets are derived from the FLoRes-101 collection\footnote{\url{https://huggingface.co/datasets/gsarti/flores_101}}, which comprises 3,001 Wikipedia sentences. The authors translated development and training sentences from FLoRes-101 into 102 languages and annotated them for ASR applications. We extracted the Bangla test dataset, which includes 920 audio files totaling 3.43 hours. Fleur's dataset consists of 3,010 training, 920 testing, and 402 validation audio files. We isolated the test files to evaluate them using our chosen models.

Common Voice is a comprehensive, multilingual ASR dataset. As of now, the dataset features 17,689 validated hours across 108 languages, with continual additions of new voices and languages \cite{ardila2019common}. The Common Voice 13 dataset includes 20.7k training, 9.23k testing, and 9.23k validation audio files. We also segregated the test files from this dataset for evaluation with our selected models.

The OpenSLR Bangla dataset, identified as OpenSLR-53, is a substantial ASR corpus sponsored by Google. It consists of a total of 232,537 recordings, amounting to 229 hours of audio data. For our evaluation purposes, we downloaded specific portions of this dataset and randomly selected 10,142 files, amounting to 10 hours of audio data.

% Besides, several other research work include \cite{hasan2019towards,rakib2023bangla}.

% Our dataset contains a total of 19.8k hours of audio whereas Fleurs, Common Voice 13, OpenSLR, and BengaliAI contain 15.61 hours, 65.71 hours, 229 hours, and 12k hours respectively. 
% While previous works have made significant contributions to speech corpus development, there is a limited number of studies specifically addressing the creation of speech corpora for the Bangla language. Therefore, our research aims to fill this gap by focusing on the automatic preparation of a large-scale Bangla speech corpus using existing audio and text data.

Our introduced dataset surpasses all other available Bangla ASR datasets in terms of dataset size and annotation strategy, as outlined in Table \ref{tab:datasets_comparison}. 
% A key distinction for our dataset is the unique annotation strategy, which sets us apart from most of the other datasets.
% , with the exception of the Bangla Speech Corpus.
Compared to other methodologies, our data annotation pipeline is specialized in several crucial aspects. First, we focus on the manual curation of channels, allowing us to select content from reputable sources, thus enhancing both relevance and diversity. Second, our pipeline leverages both Hybrid ASR and Conformer ASR Models, which are potentially fine-tuned for Bangla, resulting in more accurate transcriptions. Finally, we have implemented a duplicate removal system to remove redundant content. These features make our data annotation process an excellent fit for applications that demand high-quality, domain-specific Bangla language resources.

%% file: sections/dataset.tex
\section{Dataset}
\label{sec:dataset}

\subsection{Data Collection}
\label{ssec:data_collection}

To develop a large-scale dataset focused on diverse domains, we selected YouTube as our data source due to its extensive coverage of Bangla speech. We gathered content from popular news channels such as ATN News, Banglavision News, ZEE 24 Ghanta, News18bangla, Republic Bangla, DD Bangla News, ABP Ananda, NTV News, DBC News, BBC News Bangla, Channel 24, mytvbd news, News24, and Channel I News, among others. Additionally, we included talk shows like RTV Talkshow and ATN Bangla Talk Show. We have also incorporated travel VLOGs into our dataset.

% The training sets contain 18k hours of audio and most of these are collected from various  Bengali news channels as well as from some popular talk shows. 

\paragraph{Crawler:} To facilitate the collection of data from YouTube, we developed a web crawler that periodically collects videos using youtube-dl.\footnote{\url{https://github.com/ytdl-org/youtube-dl}} This crawler operates on a list of YouTube channels that we manually pre-select to ensure domain diversity. The crawler then lists all available videos from each channel and proceeds to download them. The download module within the crawler stores the downloaded videos in a Google Cloud Storage (GCS) bucket. The resulting collection consists of $\sim$53K hours with 42K number of videos.

% In order to facilitate Automatic Speech Recognition (ASR) training, the collection of Bengali YouTube videos is essential. To achieve this, we developed a crawler that was utilized to periodically gather YouTube videos. A user manually adds a Bangla YouTube channel to be crawled. This can be accomplished either through an HTTP API or by executing a script that triggers the video discovery service. The video discovery service is responsible for identifying all the available videos on the specified channel. It then generates a download task for each video found. The video downloader workers are assigned the download tasks and initiate the video download process for each task. Once the download is complete, the video is uploaded to a Google Cloud Storage (GCS) bucket.

\begin{figure}
    \centering
    \includegraphics[width=0.45\textwidth]{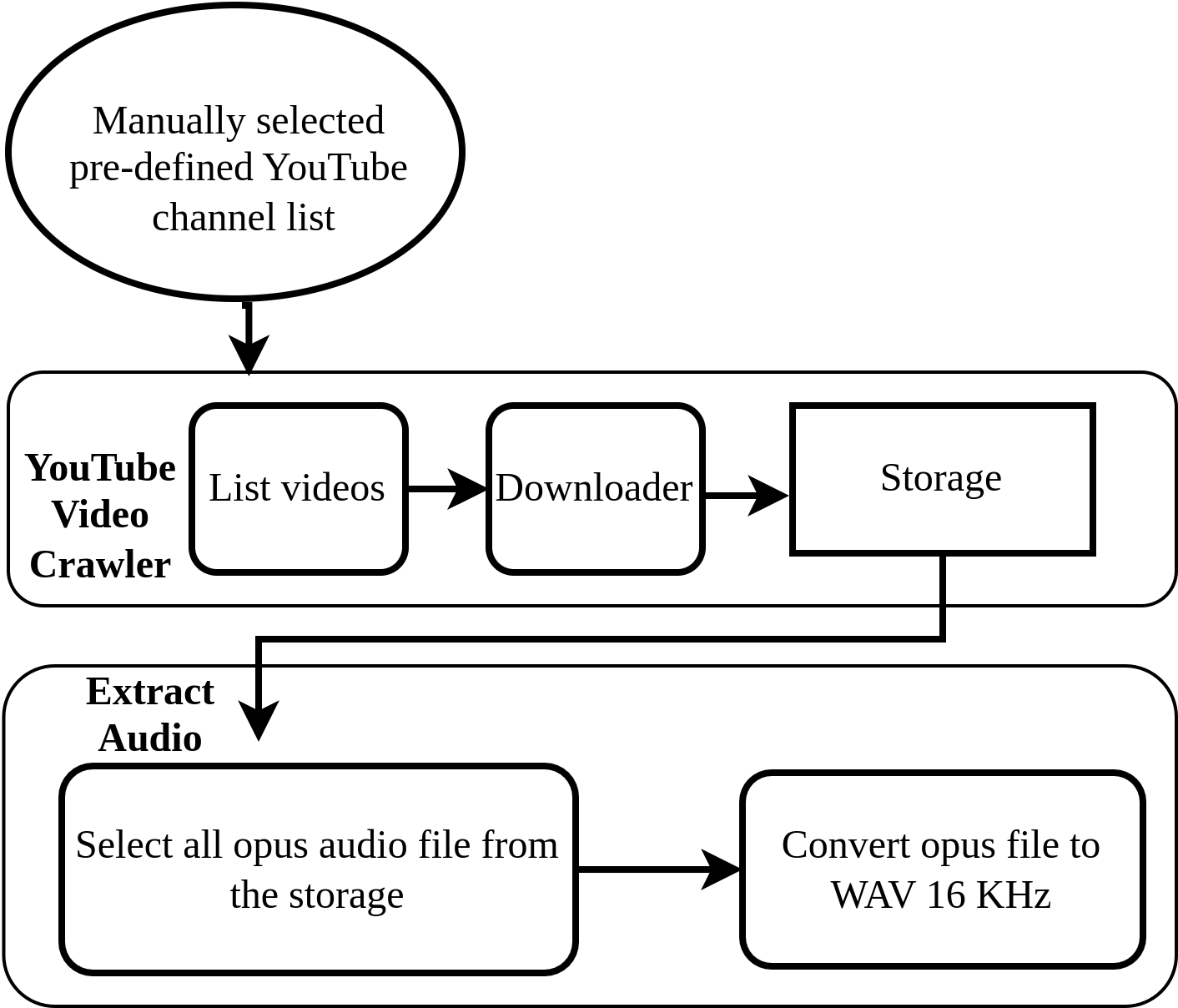}
    \caption{Data collection pipeline.}
    \label{fig:data_collection}
\end{figure}

% \paragraph{Audio Extraction from Videos}
\paragraph{Audio Extraction:} We extracted audio from the videos, which were originally in Opus format. To ensure compatibility and standardization, we converted these Opus files to WAV format with a sampling rate of 16 kHz. The conversion process demanded the use of both high CPU and low memory resources. 
% This process resulted in a total of 482,556 segments. 
In Figure \ref{fig:data_collection}, we provide the data collection pipeline.

% Next, we proceeded to extract the desired audio files from the GCS bucket, specifically targeting opus audio files. 
% These files were randomly selected and downloaded for subsequent steps. 
% Once the files were converted, they were passed through two distinct ASR systems: Vosk ASR and Nemo ASR. 

\subsection{Pseudo Labeling}
\label{ssec:data_labeling}

\begin{figure*}
    \centering
    \includegraphics[width=1.0\textwidth]{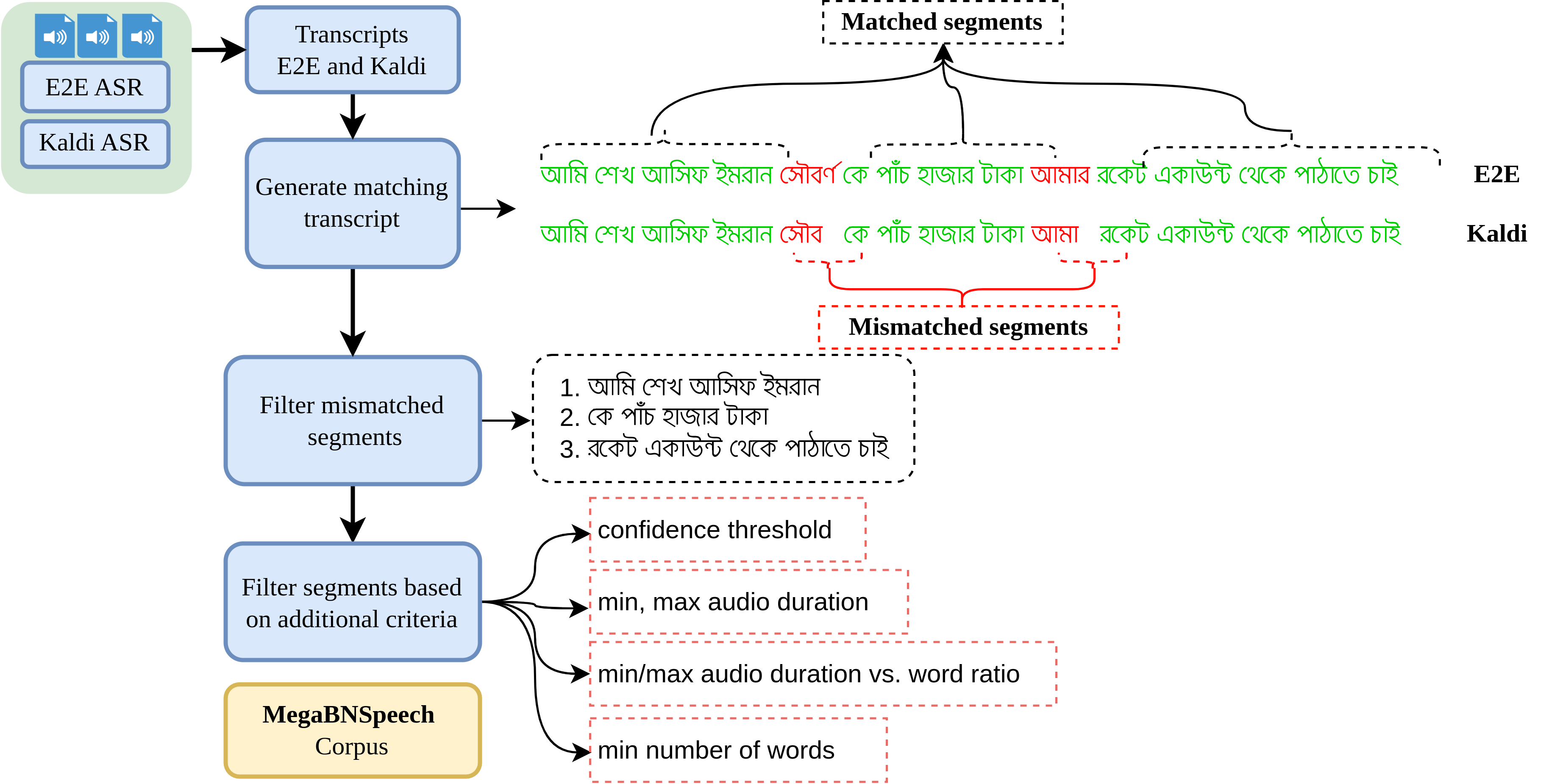}
    \caption{Architecture of the proposed pseudo labeling approach.}
    \label{fig:data_labelling}
\end{figure*}

In Figure \ref{fig:data_labelling}, we report the architecture of our proposed pseudo labeling approach for the \textit{\textbf{MegaBNSpeech}} corpus development. The system takes audio files extracted from videos and passes them into two distinct in-house developed ASR systems: 
\begin{itemize}
    \item \textbf{Hybrid ASR ($E_1$):} Kaldi \cite{povey2011kaldi} based Factorized Time Delayed Neural Network (TDNN) \cite{povey2018semi} model is used for training on 1.2K hours  transcribed YouTube audio dataset which is manually collected. The model is called hybrid because firstly a Gaussian Mixture Model (GMM) is trained on speech acoustic features for phoneme level alignment and then DNN model is trained on the aligned features. During training, we use 15 factorized TDNN layers in model architecture and 4 epochs. The training recipe is available in the Kaldi Website.\footnote{\url{https://github.com/kaldi-asr/kaldi/blob/master/egs/librispeech/s5/local/chain/tuning/run_tdnn_1d.sh}}  
    
    \item  \textbf{End-to-End Conformer ASR ($E_2$):} Nemo Tooklit based
    Conformer-CTC model \cite{gulati2020conformer} is trained on 4k hours of transcribed YouTube data. A byte-pair encoding (BPE) tokenizer  \cite{wang2005matbn} is first built using the transcripts of the train set. At training time, pretrained weights of Nemo English ASR \footnote{\url{https://catalog.ngc.nvidia.com/orgs/nvidia/teams/nemo/models/stt_en_conformer_ctc_medium}} are used for initializing weights of the encoder part only. The training parameters are epochs 16, batch size 32,  sampling rate 16kHz,  use\_start\_end\_token TRUE, pin memory TRUE, number\_of\_workers 48, trim\_silence False,  max duration 18.5 and  min duration 0.2. The training script is customized from the following the script. \footnote{\url{https://github.com/NVIDIA/NeMo/blob/main/examples/asr/conf/conformer/conformer_ctc_bpe.yaml}}   
    % \cite{}
    % 4K hours 
\end{itemize}

\noindent The objective was to leverage the capabilities of these ASR systems to generate transcription based on their decisions. We use the term expert to refer to these systems.
% matching transcripts for the audio content. This involved leveraging the capabilities of these ASR systems to convert spoken words into written text.

As part of our proposed pseudo-labeling approach, we consider them as expert systems. Based on the transcripts they generate, we take their decisions on segments that match, as depicted in Figure \ref{fig:data_labelling}. To formally define this, we have two expert systems $E_1$ and $E_2$, each of which generates transcripts $T1$ and $T2$, respectively. 
% a segment sequence $s^1, s^2...s^n$. 
% \textcolor{blue}{
We use a matching algorithm, Algorithm \ref{algo:transcription_matching},
% function $f(E_1,E_2)$ 
that employs exact string matching to align the text of segments from the experts $E_1$ and $E_2$ ASR systems. The next step involves segmenting the audio based on matching text and removing the segments that do not match. For example, the words highlighted in red in Figure \ref{fig:data_labelling} indicate mismatched segments. We therefore remove these segments. The subsequent step is to filter out segments based on predefined criteria. These include: {\em(i)} confidence score of the ASR systems, {\em(ii)} minimum and maximum duration of the segments, {\em(iii)} the ratio of segment duration to the number of words, and {\em(iv)} the minimum number of words required in a segment. 
% During this filtering process, we noticed the presence of duplicate segments and thus removed them. 
% For identifying duplicate segments, we used \textcolor{red}{to add...}. 
These steps resulted in the final MegaBNSpeech corpus. 
% Below, we provide the pseudocode for our proposed pseudo-labeling approach.

% Following the generation of the matching transcripts, we proceeded to create small audio segments based on these transcripts. This segmentation allowed for more granular analysis and processing of the audio data. However, filtering out erroneous segments based on a predetermined threshold value was necessary. This step aimed to remove any segments that did not meet the desired quality or accuracy criteria.

% Subsequently, a Nemo manifest was created based on the filtered and clean audio data. This manifest served as a structured representation of the data, facilitating the subsequent steps in the pipeline. Additionally, any duplicate data samples were identified and removed to ensure the integrity and uniqueness of the corpus.

% Finally, the prepared data, in the form of the Nemo manifest and the associated audio segments, was deemed ready for training the ASR model. This marked the completion of our corpus creation pipeline, which involved a series of carefully orchestrated steps to gather, process, filter, and organize the audio data for optimal training of the ASR model

% Our corpus creation pipeline involved a systematic process to gather and prepare audio data for the purpose of training the ASR model.  Figure~\ref{fig:data_labelling} show the Pipeline of our model creation.

\begin{algorithm}
\caption{Transcription matching algorithm.}
\label{algo:transcription_matching}
% \scalebox{0.8}{
\begin{algorithmic}[1]
\State \( T_1 \leftarrow \) Kaldi model (E1)
\State \( T_2 \leftarrow \) Conformer CTC model (E2)
\For {each \( (t_1, t_2) \) in zip (\( T_1, T_2 \))}
  \State \( \mathcal{M} \leftarrow \) \textit{f}(\( t_1, t_2 \))
  \For {each \( m \) in \( \mathcal{M} \)}
    \State \( r_w \leftarrow \) word rate of \( m \)
    \State \( d_a \leftarrow \) segment duration of \( m \)
    \State \( c_t \leftarrow \) total characters in \( m \)
    \State \( w_t \leftarrow \) total words in \( m \)
    \If {\( r_w < r_{w,\text{min}} \) \textbf{or} \( r_w > r_{w,\text{max}} \) \textbf{or} \( d_a < d_{a,\text{min}} \) \textbf{or} \( d_a > d_{a,\text{max}} \) \textbf{or} \( c_t < c_{t,\text{min}} \) \textbf{or} \( w_t < w_{t,\text{min}} \)}
      \State continue
    \EndIf
    \State Write matched transcript and segment
    % \State Write: matching transcript audio
  \EndFor
\EndFor
\end{algorithmic}
where \( r_{w,\text{min}} \) and \( r_{w,\text{max}} \) refers to minimum and maximum word rate; 
 \( d_{a,\text{min}} \) and \( d_{a,\text{max}} \) refers to minimum and maximum segment duration;
\( c_{t,\text{min}} \) refers to minimum number of characters, and 
\( w_{t,\text{min}} \) refers to minimum number of total words; \textit{f}(\( t_1, t_2 \)) is the longest substring matching function. 
\end{algorithm}

\subsection{Metadata}
To ensure both reproducibility and transferability, we store the metadata in JSON format. This metadata includes the following key elements: {\em(i)} audio\_filepath, {\em(ii)} text, and {\em(iii)} duration. The audio\_filepath field specifies the path to the audio file, with channel information embedded in the filename. The text field contains the data generated by the pseudo-labeling pipeline, while the duration field indicates the length of the audio in seconds. The audio files have a sampling rate of 16 kHz.

% Figure \ref{fig:metadata_example_test} contains an example of single audio data metadata.

% The metadata is provided in a manifest JSON line file which contains a JSON line with each line containing three keys: audio\_filepath, text, and duration. 

% \begin{figure}
%     \centering
%     \includegraphics[width=0.47\textwidth]{assets/example_metadata_testdata.png}
%     \caption{Metadata example
%     }
%     \label{fig:metadata_example_test}
% \end{figure}

%sagor 
%json_file decription of an example  training  - rewrite 

% \subsection{Training sets}
% The training sets contain 18k hours of audio and most of these are collected from various  Bengali news channels as well as from some popular talk shows. Some popular news channels are ATN News, Banglavision News, ZEE 24 Ghanta, News18bangla, Republic Bangla, DD Bangla News, ABP Ananda, NTV News, DBC News, BBC News Bangla, Channel 24, mytvbd news, News24, Channel I News, Boishakhi tv news and so on. The talk shows are RTV Talkshow, ATN Bangla talk show. We have also collected VLOG data which includes several travel VLOGs. 

%sagor 

% \section{Creation Pipeline}
% \label{sec:task_and_data}

% In this section, we introduce the detailed creation pipeline of our HishabBNSpeech corpus, including original audio collection, audio segmentation candidate generation, and candidate calibration.

%% file: sections/experiments.tex
\section{Experiments}
\label{sec:experiments}

\subsection{Data splits}
% Description 

\paragraph{Training set} 
For training the model, the dataset we selected comprises 17.64k hours of news channel content, 688.82 hours of talk shows, 0.02 hours of vlogs, and 4.08 hours of crime shows. Table \ref{tab:category_wise_training_data_dist} provides detailed information about each category and its corresponding duration in hours.

% Table:  Training set Category Hours  [Table]
\begin{table}[h]
\centering
\begin{tabular}{lr}
\hline
\textbf{Channels Category} & \textbf{Hours}\\
\hline
News & 17,640.00\\
Talkshow & 688.82\\
Vlog & 0.02 \\
Crime Show & 4.08 \\
\textbf{Total} & \textbf{18,332.92}\\ 
\hline
\end{tabular}
\caption{Training data distribution according to channel category and hours}
\label{tab:category_wise_training_data_dist}
\end{table}

\paragraph{Development set}
To investigate the robustness of the pseudo-labeling approach, we randomly selected 10 hours of speech to create a development set. 

\paragraph{Test set}
% We separate 1.8k hours of audio for evaluation and testing purposes. 
To evaluate the performance of the models, we used four test sets. Two of these were developed as part of the MegaBNSpeech corpus, while the remaining two (Fleurs and Common Voice) are commonly used test sets that are widely recognized by the speech community. 

% Among the five test sets, we prepared two test sets: 1) Hishab Youtube-Testset of 8.05 hours and 2) Telephony Testset of 1.9 hours. The other three test data are 1) Fleurs test set of 3.43 hours, 2) Common Voice test set of 16.5 hours, and 3) OpenSLR Random Set of 8 hours. The test datasets are quite diverse, with domain variants and well-suited for model performance analysis, benchmarking, and interpretation. 

\begin{itemize}
     \item \textbf{MegaBNSpeech-YT Test Set :}
    The test set has been prepared from a recent collection of YouTube videos, resulting in 8 hours of data. This set is manually transcribed for evaluation purposes. The domains of this set include News, Talkshow, Courses, Drama, Science, Waz (Islamic preaching), etc.
    \item \textbf{MegaBNSpeech-Tele Test Set:} To assess the model's generalization capabilities, we also included 1.9 hours of telephony conversations from our in-house dataset collection, which were subsequently manually transcribed. It involves telephone conversations covering various discussion topics, including online food orders, health services, online ticket bookings, and online banking. The calls were originally recorded using 8kHz sampling rate, which we then upsampled to 16kHz to match the ASR input.\footnote{The curated telephony dataset is open-ended conversations with pre-defined topics and includes consent from the interlocutors.}
    % such as \textcolor{red}{ any idea of the topics?}. 
    % and is carefully annotated. The reason for considering the Telephony test set is to introduce a different source of test set from our training data for better understanding and comparison among selected models. 
    
    \item \textbf{Fleurs:} Fleur's \cite{conneau2023fleurs} datasets are from FLoRes-101 datasets\footnote{\url{https://huggingface.co/datasets/gsarti/flores_101}} which contain 3001 Wikipedia sentences. The authors translated dev and train sentences from FLoRes-101 to 102 languages and annotated them to develop ASR. We have separated the Bangla test datasets which contain 920 audio files with 3.43 hours of data. Fleurs contains a total of 3,010 train, 920 test, and 402 validation audio files. We separated the test datasets and evaluated them with our selected models.
    \item \textbf{Common Voice:} Common voice \cite{ardila2019common} is a massively multilingual ASR dataset. The dataset currently consists of 17,689 validated hours in 108 languages, but more voices and languages are always added. Common Voice 13 contains a total of 20.7k train, 9.23k of test, and 9.23k of validation audio files. We separated the test datasets and evaluated them with our selected models.
\end{itemize}

\subsection{Contemporary ASR Models}

% The candidate models are: 1) Hishab Baseline, 2) Google ASR API, 3)MMS \cite{pratap2023scaling}, and 4) ODD-speech Conformer \cite{rakib2023ood}. MMS is a multi-lingual ASR developed by Meta and the ODD-speech ASR Conformer model is trained on the ODD speech\cite{rakib2023bangla} 11.8k dataset. 

\paragraph{Google:} Google speech-to-text\footnote{\url{https://cloud.google.com/speech-to-text}} is a cloud-based ASR service that provides transcription from input Audio for several languages. It provides different domain-specific models for task-specific ASR services. We used the default model and settings and set the language to Bangla.

\paragraph{MMS:} Massively Multilingual Speech (MMS \cite{pratap2023scaling}) is a fine-tuned model developed by Meta. This model is based on the Wav2Vec2 \cite{baevski2020wav2vec} architecture and makes use of adapter models to transcribe 1000+ languages. The model consists of 1 billion parameters and has been fine-tuned in 1,162 languages. The model checkpoint is published in the HuggingFace model hub.\footnote{\url{https://huggingface.co/facebook/mms-1b-all}}

\paragraph{OOD-speech ASR:}
OOD-speech ASR is a Conformer-CTC-based model trained on ODD speech datasets \cite{rakib2023bangla}. The model consists of 121 million parameters and is trained on 1,100+ hours of audio data which is crowdsourced from native Bangla speakers. The model was trained using NVIDIA NeMo\footnote{\url{https://docs.nvidia.com/deeplearning/nemo/user-guide/docs/en/stable/asr/examples/kinyarwanda_asr.html}} framework and published in Huggingface model hub.\footnote{\url{https://huggingface.co/bengaliAI/BanglaConformer}}

% \subsection{Data Preprocessing}

\subsection{MegaBNSpeech ASR}

% \paragraph{Settings:} 
% We have done an extensive baseline experiment with 
We trained the FastConformer model \cite{rekesh2023fast} using the full 18k MegaBNSpeech training sets. During the training phase, we employed a set of predefined parameters: a learning rate of 0.5, a weight decay of 0.001, a batch size of 32, AdamW optimizer, and a maximum audio duration of 15 seconds. We provide details of the hyperparameter settings in Table~\ref{tab:training_params}.

% for training.
% and 10 hours of audio for validation. The validation set is randomly selected from 1.8k testing sets. 

To optimize the performance of our model, we conducted experiments with various NVIDIA NeMo architectures and assessed their training accuracy. Specifically, we evaluated the Conformer-CTC, Conformer-Transducer, and Fast-Conformer models. Among these, the Conformer-CTC model exhibited the best performance, achieving a training loss of approximately 11.2\%.

To accelerate the training process, we deployed a total of 16 $A100-40G$ GPUs to handle the entire dataset. Despite leveraging significant computational resources, the training still took approximately $112$ hours to complete.

% \subsection{Training}
% \label{sss:training}
% \subsection{Training Parameters}
% We trained the model using the AdamW optimizer for a duration of 4 days, 18 hours, and 16 minutes. The model was trained on a GPU cluster consisting of 16 GPUs, enabling efficient parallel processing. 
% This configuration allowed for the exploration of large-scale datasets and contributed to the model's robustness and accuracy. Table~\ref{tab:training_params} describes the key parameters of our model evaluation.

%Need to insert a table%

% \subsection{Evaluation Measures}

% epoch
% 15
% global_step
% 90,911.296875
% learning_rate
% 0.00007328778156079
% train_backward_timing
% 0.16302824020385742
% train_loss
% 11.203718185424805
% train_step_timing
% 1.630632400512695
% global_step
% 90,910
% training_batch_wer
% 0.14923124015331268
% val_loss
% 27.589672088623047
% val_wer
% 0.20338508486747744
% validation_step_timing
% 0.08939909934997559

\begin{table}
\centering
\begin{tabular}{lc}
\hline
\textbf{Parameter} & \textbf{Value}\\
\hline
epoch & 15\\
global\_step & 90,911 \\
learning\_rate & 0.000073287 \\ 
train\_backward\_timing & 0.1630282 \\ 
train\_loss & 11.203718 \\
training\_batch\_wer & 0.149231  \\ 
val\_loss & 27.58967 \\ 
val\_wer & 0.203385 \\
validation\_step\_timing & 0.089399 \\\hline\end{tabular}
\caption{Details of the hyperparameter settings.
% Training parameters details for baseline experiment using Fast Conformer model on 18k ASR dataset
}
\label{tab:training_params}
\end{table}

% The output parameters offer valuable insights into the performance of the ASR-trained model. 
The model underwent training for 15 epochs, completing approximately 90,911 global steps. The chosen learning rate was relatively low, contributing to stable and incremental updates of the model's parameters. Although the training loss suggests potential for further improvement, it does indicate a narrowing gap between predicted and actual values during the training phase.

% As for the WER value, it can be said that our model performed with good accuracy. However, the validation loss is relatively a little high. These parameters provide valuable insights into the model's performance and can guide future optimization efforts to enhance its accuracy and reduce errors. 

As for the WER the value indicates that our model performed with commendable accuracy. However, the validation loss remains somewhat elevated. These metrics offer valuable insights into the model's performance and serve as a road map for future optimization efforts.
% aimed at enhancing accuracy and minimizing errors.

% Here are some graphs that were occupied from the training output: (?)

\subsection{Data Post-processing}
During the evaluation of the test sets, we apply a set of post-processing on predicted transcription and human annotation to reduce unexpected symbols, confused words, and misleading alignment. We find that there are some typing issues during manual labeling.
% such as \textbf{ookar} is written as \textbf{eekar} $+$ \textbf{aakar}, \textbf{oowkar} is written as \textbf{eekar} $+$ \textbf{ohikar} and so on. 
To resolve this, a typing error minimization function is applied. In addition, we added two common normalization rules including: (i) number-to-word conversation and (ii) punctuation removal. 
% There is a valid reason to use punctuation removal as we find that the word alignment is misleading for some transcriptions. 

\begin{figure} [!ht]
\centering
\includegraphics[width=0.2\textwidth]{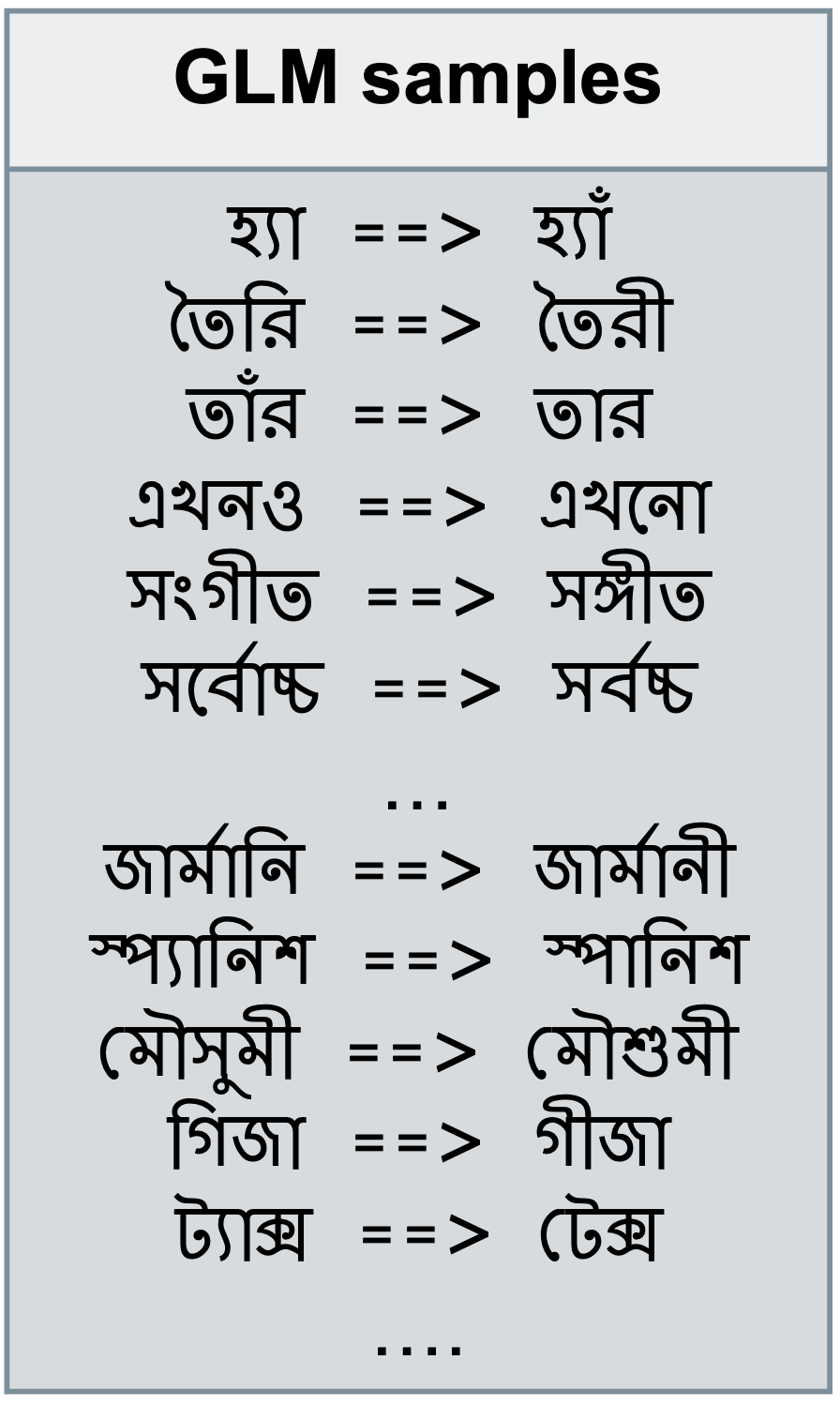}
\caption{Sample of GLM entries.}
\label{fig:glm}
\end{figure}

\paragraph{Minimizing the confusion due to writing style}
An extensive analysis of transcriptions indicates many words have different forms of writing (as shown in Figure \ref{fig:glm}) based on different character combinations. 
% is slightly different for the paired confused words. 
In some cases, both words of confused pairs are acceptable as people annotated in different ways, especially for country names, along with borrowed or code-mixed words.
% pronounced words from other language. 
% The evidence we found on google prediction on our annotated telephony and YouTube testset. 

To minimize these differences, we created a simple Global Mapping File (GLM) that allows different variations of the word to be accepted during evaluation. The GLM file contains entries for different homophones, primarily those with spelling variations.
% ways of writing a same word or proper nouns. 
We employed the most frequently occurring confusion patterns for the task, although this approach may not cover all possible variations.
% from \footnote{https://github.com/usnistgov/SCTK} toolkit. The confused word pairs which have similar pronunciation and sometimes ins translated form,  are added to the GLM file and ignored during WER Calculation.  

\subsection{Evaluation Metrics}
To evaluate the performance of the models, we used widely accepted metrics such as Word Error Rate (WER) and Character Error Rate (CER). The reported WER values are presented using the GLM and postprocessing of the hypothesis and references. 

\begin{table*} [!ht]
\centering
\begin{tabular}{llllll}
\hline
\textbf{Category} & \textbf{Duration(hr)}  & \textbf{MegaBNSpeech}  & \textbf{Google}  & \textbf{MMS} & \textbf{OOD-speech}\\
\hline
MegaBNSpeech-YT & 8.1 & 6.4/3.39&	28.3/18.88&	51.1/23.49&	44.4/33.43 \\
MegaBNSpeech-Tel & 1.9 & $^*$40.7/24.38	& $^*$59/41.26&	$^*$76.8/39.36&	$^*$69.9/52.93 \\
Fleurs & 3.42 & $^*$36.1/8.43	&24.6/8.54&	$^*$39.4/11.58&	29.5/13.97 \\
Common Voice & 16.5  & $^*$42.3/11.44 &	23.6/ 8.31&	$^*$48/14.72 & 23.6/10.49 \\
\hline
\end{tabular}
\caption{Reported Word error rate (WER) /character error rate (CER) on four test sets using four ASR systems. $^*$ represent the training portion of the corresponding test set \textbf{was not} present in the ASR model. % and $^+$ means not sure about training data.
}
\label{overall_evaluation_all_model}
\end{table*}

%% file: sections/results_and_discussion.tex
% News	2.5/1.21	18.9/10.46	52.2/21.65	32.3/20.71
% Talkshow	6/3.29	28/18.71	48.8/21.5	45.8/34.59
% Courses	6.8/3.79	30.8/21.64	50.2/23.52	46/35.99
% Drama	10.3/7.47	37.3/27.43	64.3/32.74	53.6/45.14
% Science	5/1.92	20.6/11.4	45.3/19.93	33.4/23.11
% Vlog	11.3/6.69	33/22.9	57.9/27.18	49.3/37.22
% Recpie	7.5/3.29	26.4/16.6	53.3/26.89	41.2/29.39
% Waz	9.6/5.45	33.3/23.1	57.3/27.46	59.9/50.38
% Movie	8/4.64	35.2/23.88	64.4/34.96	50.9/42.13
\begin{table*}[!ht]
\centering
\begin{tabular}{llllll}
\hline
\textbf{Category} & \textbf{Duration(hr)}  & \textbf{MegaBNSpeech ASR}  & \textbf{Google ASR}  & \textbf{MMS ASR} & \textbf{OOD-speech}\\
\hline
News & 1.21 & 2.5/1.21 & 18.9/10.46  & 52.2/21.65& 32.3/20.71 \\
Talkshow & 1.39 & 6/3.29 &	28/18.71&	48.8/21.5&	45.8/34.59 \\
Courses & 3.81 & 6.8/3.79&	30.8/21.64&	50.2/23.52&	46/35.99 \\
Drama & 0.03  & 10.3/7.47 &	37.3/27.43 &	64.3/32.74	&53.6/45.14 \\
Science & 0.26  & 5/1.92	&20.6/11.4	&45.3/19.93	&33.4/23.11 \\
Vlog & 0.18 & 11.3/6.69	&33/22.9&	57.9/27.18&	49.3/37.22	 \\
Recipie & 0.58 & 7.5/3.29&	26.4/16.6&	53.3/26.89&	41.2/29.39 \\
Waz & 0.49  & 9.6/5.45&	33.3/23.1&	57.3/27.46&	59.9/50.38 \\
Movie & 0.1  & 8/4.64	&35.2/23.88&	64.4/34.96&	50.9/42.13 \\
\hline
\end{tabular}
\caption{ Reported Word error rate (WER) /character error rate (CER) on different categories present in MegaBNSpeech - YT test set for four different ASR systems.}
\label{category_wise_test_set_evaluation_all_model}
\end{table*}

\section{Results and Discussion}
\label{sec:results}

\subsection{Robustness of Pseudo-labelling}

We first evaluate the robustness of our annotation process for unlabeled audio by utilizing our proposed pseudo-labeling approach. To investigate the quality of these annotations, we used the development set mentioned earlier. 
% we randomly selected 10 hours of speech from the dataset and used it as the development set for this study. 
This set was subsequently annotated by a human annotator who had no prior knowledge of the ASR-generated pseudo-labels. We then computed the Word Error Rate (WER) and Character Error Rate (CER) between these pseudo-labels (serving as predictions) and the human annotations (acting as ground truth). We observed WER and CER rates of less than 3\% (specifically, 2.89\% for WER and 2.27\% for CER), thereby increasing our confidence in the reliability of the pseudo-labeled datasets.

% we randomly selected 10K hours of speech data from our corpus and 

% One general question is how accurate our pseudo-annotation strategy is. We randomly select 10k hours of audio from the devset and again. Then we calculate WER and CER by considering human labeled annotation as ground truth and psedu-annotations as prediction. the resultant WER and  CER are 2.89\%,  2.27\% respectively.

\subsection{Effectiveness of MegaBNSpeech ASR}
We initially assess the performance of MegaBNSpeech ASR, which is fully trained on a pseudo-labeled dataset, and compare its ASR performance against other systems such as Google, MMS, and OOD-speech ASRs. Utilizing our in-domain test set (MegaBNSpeech-YT), we noticed a significant performance gap; MegaBNSpeech ASR outperformed the commercial Google ASR, which itself was notably better than the rest (see Table \ref{overall_evaluation_all_model}).

One plausible explanation for MegaBNSpeech's high performance could be the nature of its training data, which is predominantly sourced from News and Talkshow segments, followed by Science content. These sources typically feature formal speaking styles and limited linguistic diversity, thereby contributing to improved performance. This hypothesis is further supported by the category-level performance data, especially within the `News' category, as indicated in Table \ref{category_wise_test_set_evaluation_all_model}.

\paragraph{Across different categories:}
In Table \ref{category_wise_test_set_evaluation_all_model}, we report the WER for each category within the MegaBNSpeech-YT test set. From the table, it is evident that all the ASRs (except MMS) perform exceptionally well in the broadcast domain, specifically in News, with MegaBNSpeech achieving nearly 98\% accuracy. In the case of talk shows -- a synchronized conversational setup -- both MegaBNSpeech and Google significantly outperform MMS and OOD-speech. This trend is observed across almost all the categories.

% The possible reason is that the test set is not so diverse from the training set. Another hypothesis is our model should be better than another model from YouTube video's subtitle generation for Bengali Videos for a wide range of categories.

% \footnote{Collected from YT with an one year gap.}
% \subsection{Comparison to  }

\subsection{Generalization Capability to unknown Dataset and Channel}

\paragraph{Dataset:} 
To understand how the model performs in unknown domains or datasets, we evaluated the four ASRs using the widely used Fleurs and Common Voice test sets. As seen in Table \ref{overall_evaluation_all_model}, MegaBNSpeech performs slightly better than MMS ASR on both Fleurs and Common Voice test sets, even though these two datasets are unfamiliar to both MMS and MegaBNSpeech ASR. On the other hand, Google and OOD-speech perform significantly well, with a Word Error Rate (WER) in the range of 23-29\%. It should be noted, however, that OOD-speech ASR has been trained on Common Voice data -- a crowdsourced dataset where the text prompts are randomly selected from Wikipedia, making it similar to Fleurs. Therefore, the content and style of these datasets are not entirely unknown to these models.

\paragraph{Telephony Channel:} 
To assess how ASR models perform not just in unfamiliar domains but also across different communication channels,\footnote{The collected data was upsampled from an 8K to a 16K sampling rate to match the input sampling rates of the models.} we evaluated these four models using telephony conversational data, as shown in MegaBNSpeech-Tel Table \ref{overall_evaluation_all_model}. Our results indicate that MegaBNSpeech ASR significantly outperforms all other ASRs, with Google coming in second place. This level of performance is consistent with our earlier observations that MegaBNSpeech ASR excels in conversation-style categories like talk shows and vlogs.

\subsection{Key Points: Psuedo-labelling based ASR vs Fully-supervised ASR}
Traditional ASR training relies heavily on extensive labeled datasets, a requirement that becomes both challenging and expensive to meet for languages, dialects, and domains with limited resources. In contrast, pseudo-labeling not only enriches the training data but also diversifies domain-specific variations, as demonstrated in this study.

From our analysis, we found that MegaBNSpeech performs comparably to supervised out-of-domain (OOD) speech ASR systems, even when exposed to data or domains it has not previously encountered. This shows the efficacy of pseudo-labeling as well as the potential of both the MegaBNSpeech datasets and the model. In this study, we trained MegaBNSpeech exclusively with pseudo-labels to demonstrate the impact of this automated labeling technique. In practical applications, supplementing pseudo-labels with a small amount of manually annotated data can further enhance ASR performance while leveraging the model's strong generalization capabilities.